\documentclass[10pt,twocolumn,letterpaper]{article}

\usepackage[pagenumbers]{cvpr} 

\usepackage{graphicx}
\usepackage{amsmath}
\usepackage{amssymb}
\usepackage{booktabs}
\usepackage{xcolor}
\usepackage{soul}
\usepackage{bm}

\usepackage{times}
\usepackage{epsfig}
\usepackage{graphicx}
\usepackage{tabularx}
\usepackage{todo}
\usepackage{enumitem}
\usepackage{caption}
\usepackage{subcaption}
\usepackage{float}

\usepackage[export]{adjustbox}
\usepackage{xr}

\usepackage{tabularx,colortbl}

\makeatletter
\newcommand*{\addFileDependency}[1]{
  \typeout{(#1)}
  \@addtofilelist{#1}
  \IfFileExists{#1}{}{\typeout{No file #1.}}
}
\makeatother

\usepackage[pagebackref,breaklinks,colorlinks]{hyperref}

\usepackage[capitalize]{cleveref}
\crefname{section}{Sec.}{Secs.}
\Crefname{section}{Section}{Sections}
\Crefname{table}{Table}{Tables}
\crefname{table}{Tab.}{Tabs.}

\def\cvprPaperID{787} 
\def\confName{CVPR}
\def\confYear{2022}

\begin{document}


\title{Adversarial Attacks against a Satellite-borne Multispectral Cloud Detector}

\author{Andrew Du$^\dagger$ \quad Yee Wei Law$^\ddagger$ \quad Michele Sasdelli$^\dagger$ \quad Bo Chen$^\dagger$ \quad Ken Clarke$^\dagger$ \\ \quad Michael Brown$^\S$ \quad Tat-Jun Chin$^\dagger$  \\ 
$^\dagger$The University of Adelaide \quad $^\ddagger$University of South Australia \quad $^\S$York University}
\maketitle

\begin{abstract}
Data collected by Earth-observing (EO) satellites are often afflicted by cloud cover. Detecting the presence of clouds---which is increasingly done using deep learning---is crucial preprocessing in EO applications. In fact, advanced EO satellites perform deep learning-based cloud detection on board the satellites and downlink only clear-sky data to save precious bandwidth. In this paper, we highlight the vulnerability of deep learning-based cloud detection towards adversarial attacks. By optimising an adversarial pattern and superimposing it into a cloudless scene, we bias the neural network into detecting clouds in the scene. Since the input spectra of cloud detectors include the non-visible bands, we generated our attacks in the multispectral domain. This opens up the potential of multi-objective attacks, specifically, adversarial biasing in the cloud-sensitive bands and visual camouflage in the visible bands. We also investigated mitigation strategies against the adversarial attacks. We hope our work further builds awareness of the potential of adversarial attacks in the EO community.
\end{abstract}

\vspace{-0.5em}
\section{Introduction}\label{sec:intro}


Space provides a useful vantage point for monitoring large-scale trends on the surface of the Earth~\cite{manfreda2018use,albert2017using,yeh2020using}. Accordingly, numerous EO satellite missions have been launched or are being planned. Many EO satellites carry multispectral or hyperspectral sensors that measure the electromagnetic radiations emitted or reflected from the surface, which are then processed to form \emph{data cubes}. The data cubes are the valuable inputs to the EO applications.

However, two thirds of the surface of the Earth is under cloud cover at any given point in time~\cite{jeppesen2019cloud}. In many EO applications, the clouds occlude the targets of interest and reduce the value of the data. In fact, many weather prediction tasks actually require clear-sky measurements~\cite{liu2020hyperspectral}. Dealing with cloud cover is part-and-parcel of practical EO processing pipelines~\cite{transon2018survey, li2019deep-ieee, paoletti2019deep, mahajan2020cloud, yuan2021review}. Cloud mitigation strategies include segmenting and masking out the portion of the data that is affected by clouds~\cite{griffin2003cloud,gomez-chova2007cloud}, and restoring the cloud-affected regions~\cite{li2019cloud,meraner2020cloud,zi2021thin} as a form of data enhancement. Increasingly, deep learning forms the basis of the cloud mitigation routines~\cite{li2019deep-ieee,castelluccio2015land,sun2020satellite,yang2019cdnet}.


\begin{figure}[t]
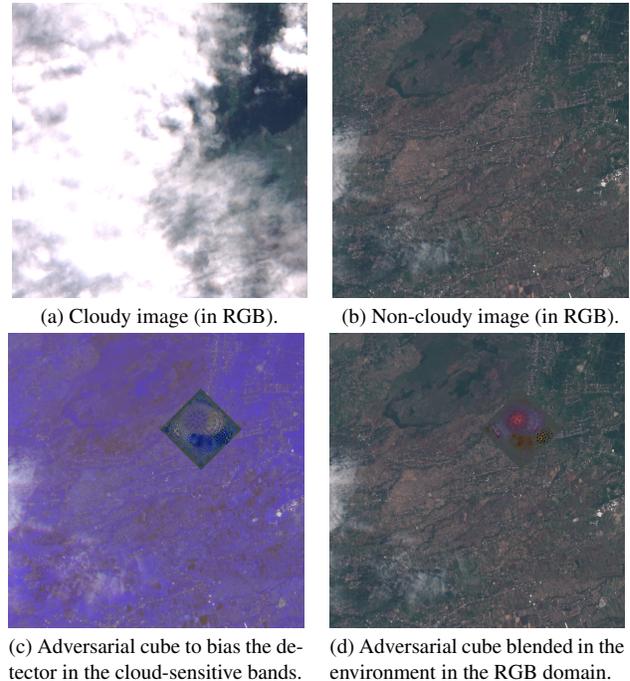
\centering
    \begin{subfigure}[b]{0.47\linewidth}
        \centering
        \includegraphics[width=\linewidth]{./figures/intro/rgb_cloudy.pdf}
        \caption{Cloudy image (in RGB).}
    \end{subfigure}
    \hspace{0.5em}
    \begin{subfigure}[b]{0.47\linewidth}
        \centering
        \includegraphics[width=\linewidth]{./figures/intro/rgb_notcloudy.pdf}
        \caption{Non-cloudy image (in RGB).}
    \end{subfigure}
    \begin{subfigure}[b]{0.47\linewidth}
        \centering
        \includegraphics[width=\linewidth]{./figures/intro/b128_patch.pdf}
        \caption{Adversarial cube to bias the detector in the cloud-sensitive bands.}
        \label{fig:falsecolor}
    \end{subfigure}
    \hspace{0.5em}
    \begin{subfigure}[b]{0.47\linewidth}
        \centering
        \includegraphics[width=\linewidth]{./figures/intro/rgb_patch.pdf}
        \caption{Adversarial cube blended in the environment in the RGB domain.}
    \end{subfigure}
\vspace{-0.5em}
\caption{(Row 1) Cloudy and non-cloudy scenes. (Row 2) Our \emph{adversarial cube} fools the multispectral cloud detector~\cite{giuffrida2020cloudscout} to label the non-cloudy scene as cloudy with high confidence.}
\label{fig:example}
\end{figure}

As the onboard compute capabilities of satellites improve, it has become feasible to conduct cloud mitigation directly on the satellites~\cite{li2018onboard,giuffrida2020cloudscout}. A notable example is CloudScout~\cite{giuffrida2020cloudscout}, which was tailored for the PhiSat-1 mission~\cite{esa-phisat-1} of the European Space Agency (ESA). PhiSat-1 carries the HyperScout-2 imager~\cite{esposito2019in-orbit} and the Eyes of Things compute payload~\cite{deniz2017eyes}. Based on the multispectral measurements, a convolutional neural network (CNN) is executed on board to perform cloud detection, which, in the case of~\cite{giuffrida2020cloudscout}, involves making a binary decision on whether the area under a data cube is \emph{cloudy} or \emph{not cloudy}; see Fig.~\ref{fig:example} (Row 1). To save bandwidth, only \emph{non-cloudy} data cubes are downlinked, while \emph{cloudy} ones are not transmitted to ground~\cite{giuffrida2020cloudscout}.

However, deep neural networks (DNNs) in general and CNNs in particular are vulnerable towards adversarial examples, \ie, carefully crafted inputs aimed at fooling the networks into making incorrect predictions~\cite{akhtar2018threat, yuan2019adversarial}. A particular class of adversarial attacks called physical attacks insert adversarial patterns into the environment that, when imaged together with the targeted scene element, can bias DNN inference~\cite{athalye2018synthesizing, brown2017adversarial, eykholt2018robust, sharif2016accessorize, thys2019fooling}. In previous works, the adversarial patterns were typically colour patches optimised by an algorithm and fabricated to conduct the attack.

It is natural to ask if DNNs for EO data are susceptible to adversarial attacks. In this paper, we answer the question in the affirmative by developing a physical adversarial attack against a multispectral cloud detector~\cite{giuffrida2020cloudscout}; see Fig.~\ref{fig:example} (Row 2). Our adversarial pattern is optimised in the multispectral domain (hence is an \emph{adversarial cube}) and can bias the cloud detector to assign a \emph{cloudy} label to a \emph{non-cloudy} scene. Under the mission specification of CloudScout~\cite{giuffrida2020cloudscout}, EO data over the area will not be transmitted to ground.

\vspace{-1em}
\paragraph{Our contributions} 

Our specific contributions are:
\begin{enumerate}[leftmargin=1em,itemsep=2pt,parsep=0pt,topsep=2pt]
\item We demonstrate the optimisation of adversarial cubes to be realised as an array of exterior paints that exhibit the multispectral reflectance to bias the cloud detector.
\item We propose a novel multi-objective adversarial attack concept, where the adversarial cube is optimised to bias the cloud detector in the cloud sensitive bands, while remaining visually camouflaged in the visible bands.
\item We investigate mitigation strategies against our adversarial attack and propose a simple robustification method.
\end{enumerate}

\vspace{-1em}
\paragraph{Potential positive and negative impacts}

Research into adversarial attacks can be misused for malicious activities. On the other hand, it is vital to highlight the potential of the attacks so as to motivate the development of mitigation strategies. Our contributions above are aimed towards the latter positive impact, particularly \#3 where a defence method is proposed. We are hopeful that our work will lead to adversarially robust DNNs for cloud detection.

\section{Related work}\label{sec:related_work}

Here, we review previous works on dealing with clouds in EO data and adversarial attacks in remote sensing.

\subsection{Cloud detection in EO data}\label{sec:related_hyperspectral}

EO satellites are normally equipped with multispectral or hyperspectral sensors, the main differences between the two being the spectral and spatial resolutions~\cite{madry2017electrooptical,transon2018survey}. Each ``capture'' by a multi/hyperspectral sensor produces a data cube, which consists of two spatial dimensions with as many channels as spectral bands in the sensor.

Since 66-70\% of the surface of the Earth is cloud-covered at any given time~\cite{jeppesen2019cloud,li2018onboard}, dealing with clouds in EO data is essential. Two major goals are:
\begin{itemize}[leftmargin=1em,itemsep=2pt,parsep=0pt,topsep=2pt]
\item Cloud detection, where typically the location and extent cloud coverage in a data cube is estimated;
\item Cloud removal~\cite{li2019cloud,meraner2020cloud,zi2021thin}, where the values in the spatial locations occluded by clouds are restored.
\end{itemize}
Since our work relates to the former category, the rest of this subsection is devoted to cloud detection.


Cloud detection assigns a \emph{cloud probability} or \emph{cloud mask} to each pixel of a data cube. The former indicates the likelihood of cloudiness at each pixel, while the latter indicates discrete levels of cloudiness at each pixel~\cite{sinergise-cloud-masks}. In the extreme case, a single binary label (\emph{cloudy} or \emph{not cloudy}) is assigned to the whole data cube~\cite{giuffrida2020cloudscout}; our work focusses on this special case of cloud detection.


Cloud detectors use either \emph{hand-crafted features} or \emph{deep features}. The latter category is of particular interest because the methods have shown state-of-the-art performance~\cite{lopezpuigdollers2021benchmarking,liu2021dcnet}. The deep features are extracted from data via a series of hierarchical layers in a DNN, where the highest-level features serve as optimal inputs (in terms of some loss function) to a classifier, enabling discrimination of subtle inter-class variations and high intra-class variations~\cite{li2019deep-ieee}. The majority of cloud detectors that use deep features are based on an extension or variation of Berkeley's fully convolutional network architecture~\cite{long2015fully, shelhamer2017fully}, which was designed for pixel-wise semantic segmentation and demands nontrivial computing resources. For example, \cite{li2019deep} is based on SegNet~\cite{badrinarayanan2017segnet}, while \cite{mohajerani2018cloud, jeppesen2019cloud, yang2019cdnet, lopezpuigdollers2021benchmarking, liu2021dcnet, zhang2021cnn} are based on U-Net~\cite{ronneberger2015u-net}, all of which are not suitable for on-board implementation.

\subsection{On-board processing for cloud detection}


On-board cloud detectors can be traced back to the thresholding-based Hyperion Cloud Cover algorithm~\cite{griffin2003cloud}, which operated on 6 of the hyperspectral bands of the EO-1 satellite. Li \etal's on-board cloud detector~\cite{li2018onboard} is an integrative application of the techniques of decision tree, spectral angle map~\cite{decarvalhojr2000spectral}, adaptive Markov random field~\cite{zhang2011adaptive} and dynamic stochastic resonance~\cite{chouhan2013enhancement}, but no experimental feasibility results were reported. Arguably the first DNN-based on-board cloud detector is CloudScout~\cite{giuffrida2020cloudscout}, which operates on the HyperScout-2 imager~\cite{esposito2019in-orbit} and Eye of Things compute payload~\cite{deniz2017eyes}. As alluded to above, the DNN assigns a single binary label to the whole input data cube; details of the DNN will be provided in Sec.~\ref{sec:training}.


\subsection{Adversarial attacks in remote sensing}



Adversarial examples can be \emph{digital} or \emph{physical}. Digital attacks apply pixel-level perturbations to legitimate test images, subject to the constraints that these perturbations look like natural occurrences, \eg, electronic noise. Classic white-box attacks such as the FGSM~\cite{goodfellow2015explaining}
have been applied to attacking CNN-based classifiers for RGB images~\cite{xu2021assessing}, multispectral images~\cite{kalin2021automating} and synthetic aperture radio images~\cite{li2021adversarial}. A key observation is the generalisability of attacks from RGB to multispectral images~\cite{ortiz2018integrated, ortiz2018on}. Generative adversarial networks have been used to generate natural-looking hyperspectral adversarial examples~\cite{burnel2021generating}.

Physical attacks, as defined in Sec.~\ref{sec:intro}, need only access to the environment imaged by the victim, whereas digital attacks need access to the victim's test images (\eg, in a memory buffer); in this sense, physical attacks have weaker operational requirements and the associated impact is more concerning. For \emph{aerial/satellite RGB imagery}, physical attacks on a classifier~\cite{czaja2018adversarial}, aircraft detectors~\cite{den2020adversarial, lu2021scale} and a car detector~\cite{du2022physical} have been investigated but only \cite{du2022physical} provided real-world physical test results. For \emph{aerial/satellite multi/hyperspectral imagery}, our work is arguably the first to consider physical adversarial attacks.

\section{Threat model}\label{sec:threat_model}

We first define the threat model that serves as a basis for our proposed adversarial attack.

\begin{description}[leftmargin=1em,itemsep=2pt,parsep=0pt,topsep=2pt]
\item[Attacker's goals] The attacker aims to generate an adversarial cube that can bias a pretrained multispectral cloud detector to label non-cloudy space-based observation of scenes on the surface as cloudy. In addition, the attacker would like to visually camouflage the cube in a specific \textbf{region of attack (ROA)}; see Fig.~\ref{fig:rgb_scenes} for examples. Finally, the cube should be physically realisable.


\begin{figure}[ht]
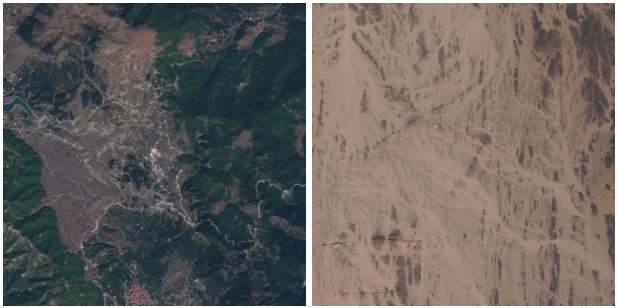
\centering
     \begin{subfigure}[b]{0.23\textwidth}
         \centering
         \includegraphics[width=\textwidth]{./figures/threat_model/hills-roa.pdf}
         \caption{Hills.}
         \label{fig:hills}
     \end{subfigure}
     \begin{subfigure}[b]{0.23\textwidth}
         \centering
         \includegraphics[width=\textwidth]{./figures/threat_model/desert-roa.pdf}
         \caption{Desert.}
         \label{fig:desert}
     \end{subfigure}
\vspace{-0.5em}
\caption{Sample regions of attack.}
\label{fig:rgb_scenes}
\end{figure}

\item[Attacker's knowledge] The attacker has full information of the targeted DNN, including architecture and parameter values, \ie, white-box attack. This is a realistic assumption due to the publication of detailed information on the model and training data~\cite{giuffrida2020cloudscout}. Moreover, from a threat mitigation viewpoint, assuming the worst case is useful.

\item[Attacker's strategy] The attacker will optimise the adversarial cube on training data sampled from the same input domain as the cloud detector; the detailed method will be presented in Sec.~\ref{sec:attacking}. The cube will then be fabricated and placed in the environment, including the ROA, although Sec.~\ref{sec:limitations} will describe limitations on real-world evaluation of the proposed attack in our study.

\end{description}


\section{Building the cloud detector}\label{sec:training}

We followed Giuffrida \etal.~\cite{giuffrida2020cloudscout} to build a multispectral cloud detector suitable for satellite deployment.

\subsection{Dataset}\label{sec:cloud_detectors}

We employed the Cloud Mask Catalogue~\cite{francis_alistair_2020_4172871}, which contains cloud masks for 513 Sentinel-2A~\cite{2021sentinel-2} data cubes collected from a variety of geographical regions, each with 13 spectral bands and 20 m ground resolution (1024$\times$1024 pixels). Following Giuffrida \etal., who also used Sentinel-2A data, we applied the Level-1C processed version of the data, \ie, top-of-atmosphere reflectance data cubes. We further spatially divide the data into 2052 data (sub)cubes of 512$\times$512 pixels each.

To train the cloud detector model, the data cubes were assigned a binary label (\textit{cloudy} vs.~\textit{not cloudy}) by thresholding the number of cloud pixels in the cloud masks. Following Giuffrida \etal., two thresholds were used: 30\%, leading to dataset version TH30, and 70\%, leading to dataset version TH70 (the rationale will be described later). Each dataset was further divided into training, validation, and testing sets. Table~\ref{tab:cm_dataset} in the supp.~material summarises the datasets.

\subsection{Model}

We employed the CNN of Giuffrida \etal., which contains four convolutional layers in the feature extraction layers and two fully connected layers in the decision layers (see Fig.~\ref{fig:cnn_model} in the supp.~material for more details). The model takes as input 3 of the 13 bands of Sentinel-2A: band 1 (coastal aerosol), band 2 (blue), and band 8 (NIR). These bands correspond to the cloud-sensitive wavelengths; see Fig.~\ref{fig:falsecolor} for a false colour image in these bands. Using only 3 bands also leads to a smaller CNN ($\le 5$ MB) which allows it to fit on the compute payload of CloudScout~\cite{giuffrida2020cloudscout}.

Calling the detector ``multispectral'' can be inaccurate given that only 3 bands are used. However, in Sec.~\ref{sec:mitigation}, we will investigate adversarial robustness by increasing the input bands and model parameters of Giuffrida \etal.'s model.

\subsection{Training}

Following~\cite{giuffrida2020cloudscout}, a two stage training process was applied:
\begin{enumerate}[leftmargin=1em,itemsep=2pt,parsep=0pt,topsep=2pt]
\item Train on TH30 to allow the feature extraction layers to recognise ``cloud shapes''.
\item Then, train on TH70 to fine-tune the decision layers, while freezing the weights in the feature extraction layers.
\end{enumerate}
The two stage training is also to compensate for unbalanced distribution of training samples. Other specifications (\eg, learning rate and decay schedule, loss function) also follow that of Giuffrida \etal.; see~\cite{giuffrida2020cloudscout} for details.

Our trained model has a memory footprint of 4.93 MB (1,292,546 32-bit float weights), and testing accuracy and false positive rate of 95.07\% and 2.46\%, respectively.

\section{Attacking the cloud detector}\label{sec:attacking}

Here, we describe our approach to optimising adversarial cubes to attack multispectral cloud detectors.

\subsection{Adversarial cube design}\label{sec:material_selection}

Digitally, an adversarial cube $\mathbf{P}$ is the tensor
\begin{equation*}
\mathbf{P} = 
\begin{pmatrix}
 \mathbf{p}_{1,1} &  \mathbf{p}_{1,2} & \cdots &  \mathbf{p}_{1,N} \\
 \mathbf{p}_{2,1} &  \mathbf{p}_{2,2} & \cdots &  \mathbf{p}_{2,N} \\
\vdots  & \vdots  & \ddots & \vdots  \\
 \mathbf{p}_{M,1} &  \mathbf{p}_{M,2} & \cdots &  \mathbf{p}_{M,N}
\end{pmatrix} \in [0,1]^{M \times N \times 13},
\end{equation*}
where $M$ and $N$ (in pixels) are the sizes of the spatial dimensions, and $\mathbf{p}_{i,j} \in [0,1]^{13}$ is the intensity at pixel $(i,j)$ corresponding to the 13 multispectral bands of Sentinel-2A.

\begin{figure}[ht]
    \centering
    \includegraphics[width=1.0\columnwidth]{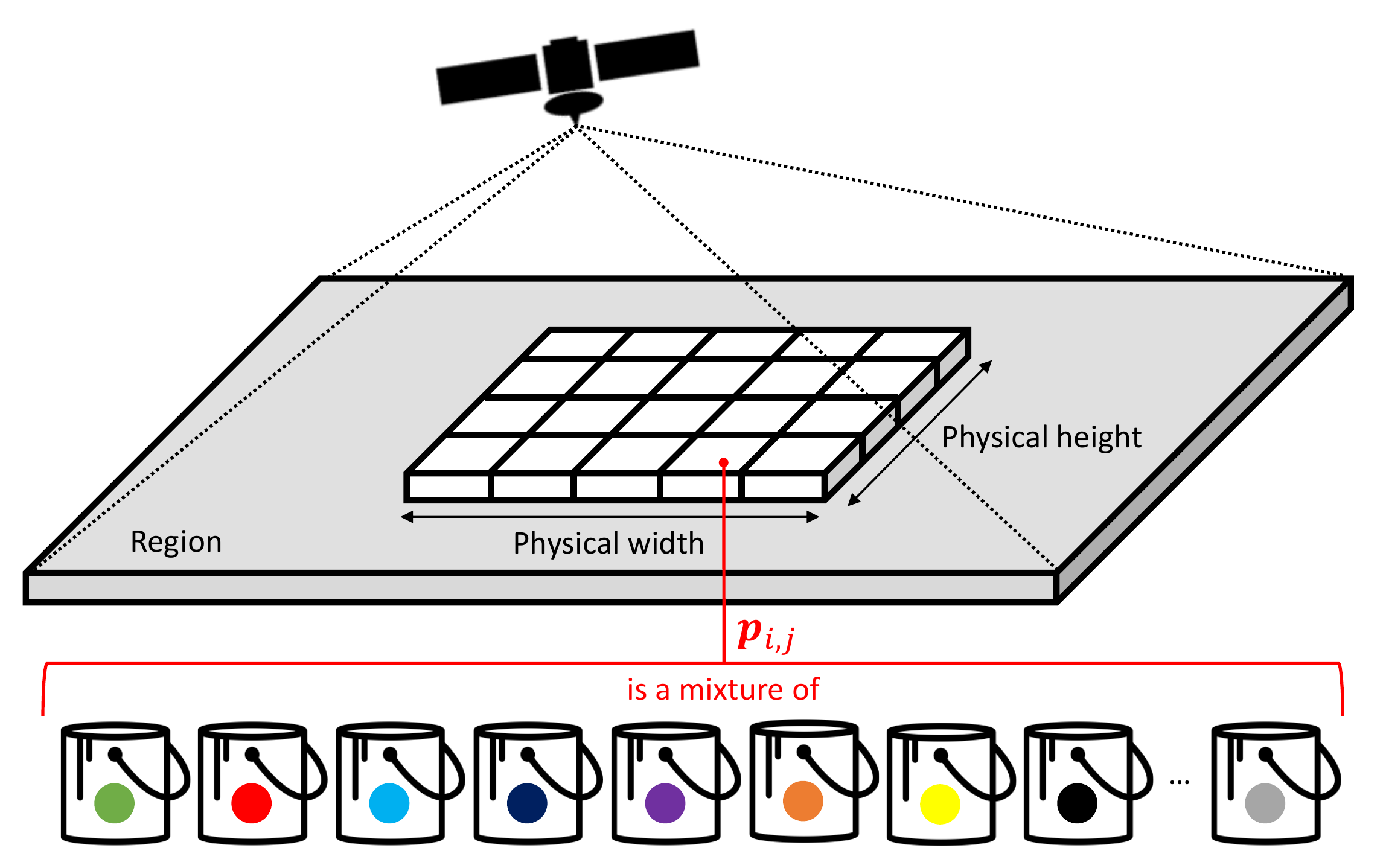}
    \vspace{-2.0em}
    \caption{The adversarial cube (digital size $4 \times 5$ pixels in the example) is to be physically realised as a mixture of exterior paint colours that generate the optimised multispectral responses.}
    \label{fig:colour_swatches}
\end{figure}

Physically, $\mathbf{P}$ is to be realised as an array of exterior paint mixtures (see Fig.~\ref{fig:colour_swatches}) that exhibit the multispectral responses to generate the attack. The real-world size of each pixel of $\mathbf{P}$ depends on the ground resolution of the satellite-borne multispectral imager (more on this in Sec.~\ref{sec:limitations}).

\subsubsection{Material selection and measurement}

\begin{figure}[ht]
    \centering
    \includegraphics[width=1.0\columnwidth]{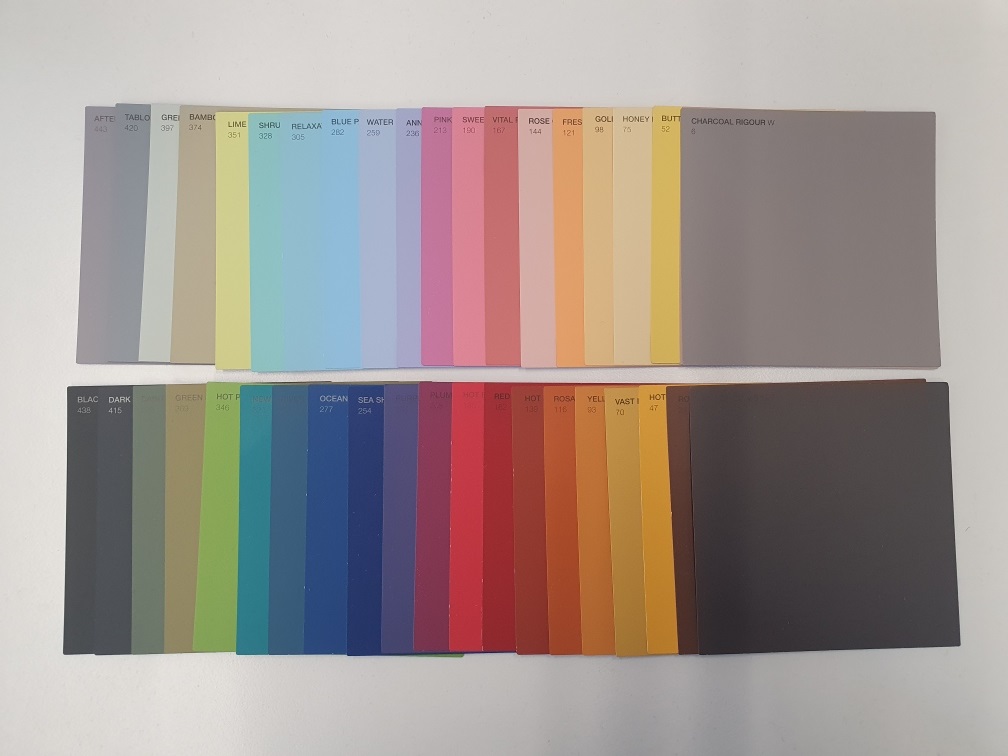}
    \vspace{-1.5em}
    \caption{A subset of our colour swatches (paint samples).}
    \label{fig:colour_swatches_real}
\end{figure}

\begin{figure*}[ht]
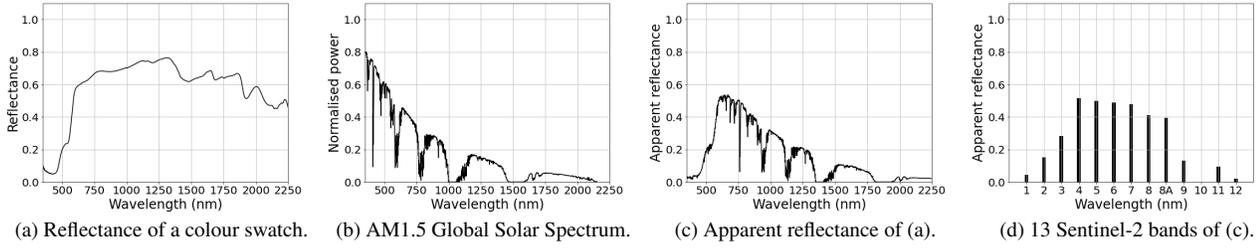
\centering
     \begin{subfigure}[b]{0.24\textwidth}
         \includegraphics[width=\textwidth]{./figures/methods/ybr_reflectance.pdf}
         \caption{Reflectance of a colour swatch.}
         \label{fig:paint_reflectance}
     \end{subfigure}
     \begin{subfigure}[b]{0.24\textwidth}
         \includegraphics[width=\textwidth]{./figures/methods/solar_spectrum.pdf}
         \caption{AM1.5 Global Solar Spectrum.}
         \label{fig:solar_spectrum}
     \end{subfigure}
     \begin{subfigure}[b]{0.24\textwidth}
         \includegraphics[width=\textwidth]{./figures/methods/ybr_apparent_reflectance.pdf}
         \caption{Apparent reflectance of (a).} 
         \label{fig:paint_apparent_reflectance}
     \end{subfigure}
     \begin{subfigure}[b]{0.24\textwidth}
         \includegraphics[width=\textwidth]{./figures/methods/ybr_13bands.pdf}
         \caption{13 Sentinel-2 bands of (c).}
         \label{fig:paint_13bands}
     \end{subfigure}
       	\vspace{-0.5em}
        \caption{Process of obtaining the 13 Sentinel-2 spectral bands of a colour swatch.}
        \label{fig:spectrometer}
\end{figure*}

To determine the appropriate paint mixtures for $\mathbf{P}$, we first build a library of multispectral responses of exterior paints. Eighty exterior paint swatches (see Fig.~\ref{fig:colour_swatches_real}) were procured and scanned with a Field Spec Pro 3 spectrometer~\cite{asd2008fieldspec3} to measure their reflectance (Fig.~\ref{fig:paint_reflectance}) under uniform illumination. To account for solar illumination when viewed from the orbit, the spectral power distribution of sunlight (specifically, using the AM1.5 Global Solar Spectrum\cite{astm2003specification}; Fig.~\ref{fig:solar_spectrum}) was factored into our paint measurements via element-wise multiplication to produce the apparent reflection; Fig.~\ref{fig:paint_apparent_reflectance}. Lastly, we converted the continuous spectral range of the apparent reflectance of a colour swatch to the 13 Sentinel-2A bands by averaging over the bandwidth of each band; Fig.~\ref{fig:paint_13bands}. The overall result is the matrix
\begin{align}
    \mathbf{C} = \left[ \begin{matrix} \mathbf{c}_1, \mathbf{c}_2, \dots, \mathbf{c}_{80} \end{matrix} \right] \in [0,1]^{13 \times 80}
\end{align}
called the \emph{spectral index}, where $\mathbf{c}_q \in [0,1]^{13}$ contains the reflectance of the $q$-th colour swatch over the 13 bands.

\subsubsection{Adversarial cube parametrisation}

We obtain $\mathbf{p}_{i,j}$ as a linear combination of the spectral index
\begin{align}\label{eq:convex}
    \mathbf{p}_{i,j} = \mathbf{C}\cdot \sigma(\mathbf{a}_{i,j}),
\end{align}
where $\mathbf{a}_{i,j}$ is the real vector
\begin{align}
\mathbf{a}_{i,j} = \left[ \begin{matrix} a_{i,j,1} & a_{i,j,2} & \dots & a_{i,j,80} \end{matrix} \right]^T \in \mathbb{R}^{80},
\end{align}
and $\sigma$ is the softmax function
\begin{align}
    \sigma(\mathbf{a}_{i,j}) = \frac{1}{\sum^{80}_{d=1} e^{a_{i,j,d}}} \left[ \begin{matrix} e^{a_{i,j,1}} & \dots & e^{a_{i,j,80}} \end{matrix} \right]^T.
\end{align}
Effectively, $\mathbf{p}_{i,j}$~\eqref{eq:convex} is a convex combination of $\mathbf{C}$.

Defining each $\mathbf{p}_{i,j}$ as a linear combination of $\mathbf{C}$ supports the physical realisation of each $\mathbf{p}_{i,j}$ through proportional mixing of the existing paints, as in colour printing~\cite{sharma2017digital}. Restricting the combination to be convex, thereby placing each $\mathbf{p}_{i,j}$ in the convex hull of $\mathbf{C}$, contributes to the sparsity of the coefficients~\cite{caratheodory-theorem}. In Sec.~\ref{sec:opimisation}, we will introduce additional constraints to further enhance physical realisability.

To enable the optimal paint mixtures to be estimated, we collect the coefficients for all $(i,j)$ into the set
\begin{align}
\mathcal{A} = \{ \mathbf{a}_{i,j} \}^{j = 1,\dots,N}_{i=1,\dots,M},
\end{align}
and parametrise the adversarial cube as
\begin{equation*}
\mathbf{P}(\mathcal{A}) = 
\begin{pmatrix}
 \mathbf{C}\sigma(\mathbf{a}_{1,1}) &  \mathbf{C}\sigma(\mathbf{a}_{1,2}) & \cdots & \mathbf{C}\sigma(\mathbf{a}_{1,N}) \\
 \mathbf{C}\sigma(\mathbf{a}_{2,1}) & \mathbf{C}\sigma(\mathbf{a}_{2,2}) & \cdots & \mathbf{C}\sigma(\mathbf{a}_{2,N}) \\
\vdots  & \vdots  & \ddots & \vdots  \\
 \mathbf{C}\sigma(\mathbf{a}_{M,1}) &  \mathbf{C}\sigma(\mathbf{a}_{M,2}) & \cdots & \mathbf{C}\sigma(\mathbf{a}_{M,N})
\end{pmatrix},
\end{equation*}
and where $\mathbf{p}_{i,j}(\mathcal{A})$ is pixel $(i,j)$ of $\mathbf{P}(\mathcal{A})$. Optimising a cube thus reduces to estimating $\mathcal{A}$.

\subsection{Data collection for cube optimisation}\label{sec:data_collection}

Based on the attacker's goals (Sec.~\ref{sec:threat_model}), we collected Sentinel-2A Level-1C data products~\cite{2021copernicus} over the globe with a distribution of surface types that resembles the Hollstein dataset~\cite{hollstein2016ready-to-use}. The downloaded data cubes were preprocessed following~\cite{francis_alistair_2020_4172871}, including spatial resampling to achieve a ground resolution of 20~m and size $512 \times 512 \times 13$. Sen2Cor~\cite{main-knorn2017sen2cor} was applied to produce probabilistic cloud masks, and a threshold of 0.35 was applied on the probabilities to decide \textit{cloudy} and \textit{not cloudy} pixels. The binary cloud masks were further thresholded with 70\% cloudiness (Sec.~\ref{sec:cloud_detectors}) to yield a single binary label for each data cube. The data cubes were then evaluated with the cloud detector trained in Sec.~\ref{sec:training}. Data cubes labelled \emph{not cloudy} by the detector was separated into training and testing sets
\begin{align}
\mathcal{D} = \{ \mathbf{D}_k \}^{2000}_{k=1}, \;\;\;\; \mathcal{E} = \{ \mathbf{E}_\ell \}^{400}_{\ell=1},
\end{align}
for adversarial cube training. One data cube $\mathbf{T} \in \mathcal{D}$ is chosen as the ROA (Sec.~\ref{sec:threat_model}).

\begin{figure*}[ht]\centering
	 \includegraphics[width=0.95\linewidth]{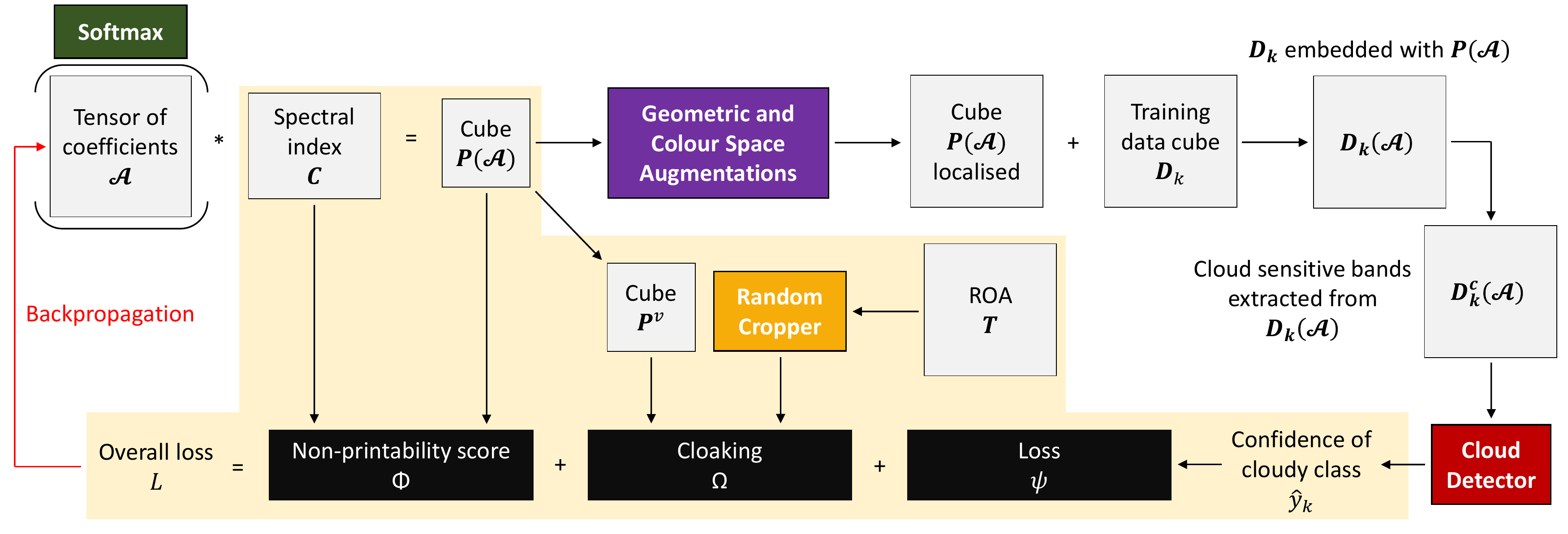} 
  	 \vspace{-0.5em}
	 \caption{Optimisation process for generating adversarial cubes.}
	 \label{fig:pipeline}
\end{figure*}

\subsection{Optimising adversarial cubes}\label{sec:patch}

We adapted Brown \etal's~\cite{brown2017adversarial} method, originally developed for optimising adversarial patches (visible domain). Fig.~\ref{fig:pipeline} summarises our pipeline for adversarial cube optimisation, with details provided in the rest of this subsection.

\vspace{-1em}
\paragraph{Subcubes}

First, we introduce the subcube notation. Let $b \subseteq \{1,2,\dots,13\}$ index a subset of the Sentinel-2A bands. Using $b$ in the superscript of a data cube, e.g., $\mathbf{P}^{b}$, implies extracting the subcube of $\mathbf{P}$ with the bands indexed by $b$. Of particular interest are the following two band subsets:
\begin{itemize}[leftmargin=1em,itemsep=2pt,parsep=0pt,topsep=2pt]
    \item $c = \{1, 2, 8\}$, \ie, the cloud sensitive bands used in~\cite{giuffrida2020cloudscout}. 
    \item $v = \{2, 3, 4\}$, \ie, the visible bands.
\end{itemize}

\subsubsection{Cube embedding and augmentations}\label{sec:augmentations}


Given the current $\mathcal{A}$, adversarial cube $\mathbf{P}(\mathcal{A})$ is embedded into a training data cube $\mathbf{D}_k$ through several geometric and spectral intensity augmentations that simulate the appearance of the adversarial cube when captured in the field by a satellite. The geometric augmentations include random rotations and positioning to simulate variations in placement of $\mathbf{P}(\mathcal{A})$ in the scene. The spectral intensity augmentations include random additive noise, scaling and corruption to simulate perturbation by ambient lighting.

\subsubsection{Loss function and optimisation}\label{sec:opimisation}

Define $\mathbf{D}_k(\mathcal{A})$ as the training data cube $\mathbf{D}_k$ embedded with $\mathbf{P}(\mathcal{A})$ (with the augmentations described in Sec.~\ref{sec:augmentations}). The data cube is forward propagated through the cloud detector $f$ to estimate the \emph{confidence}
\begin{align}
    \hat{y}_k = f(\mathbf{D}^c_k(\mathcal{A}))
\end{align}
of $\mathbf{D}_k(\mathcal{A})$ being in the \emph{cloudy} class. Note that the cloud detector considers only the subcube $\mathbf{D}^c_k(\mathcal{A})$ corresponding to the cloud sentitive bands. Since we aim to bias the detector to assign high $\hat{y}_k$ to $\mathbf{D}_k(\mathcal{A})$, we construct the loss
\begin{align}\label{eq:loss}
\Psi(\mathcal{A},\mathcal{D}) = \sum_k -\log(f(\mathbf{D}^c_k(\mathcal{A}))).
\end{align}

In addition to constraining the spectral intensities in $\mathbf{P}(\mathcal{A})$ to be in the convex hull of $\mathbf{C}$, we also introduce the multispectral non-printability score (NPS)
\begin{align}\label{eq:nps_loss}
    \Phi(\mathcal{A}, \mathbf{C}) = \frac{1}{M N} \sum_{i,j} \left( \min_{\textbf{c} \in \mathbf{C}} \left\| \textbf{p}_{i,j}(\mathcal{A})  - \mathbf{c}\right\|_2 \right).
\end{align}
Minimising $\Phi$ encourages each $\textbf{p}_{i,j}(\mathcal{A})$ to be close to (one of) the measurements in $\textbf{C}$, which sparsifies the coefficients $\sigma(\mathbf{a}_{i,j})$ and helps with the physical realisability of $\mathbf{P}(\mathcal{A})$. The multispecral NPS is an extension of the original NPS for optimising (visible domain) adversarial patches~\cite{sharif2016accessorize}.

To produce an adversarial cube that is ``cloaked'' in the visible domain in the ROA defined by $\mathbf{T}$, we devise the term
\begin{align}\label{eq:cloaking_loss}
    \Omega(\mathcal{A}, \mathbf{T}) = \left\| \textbf{P}^{v}(\mathcal{A}) - \mathbf{T}^v_{M \times N} \right\|_2,
\end{align}
where $\mathbf{T}^v_{M \times N}$ is a randomly cropped subcube of spatial height $M$ and width $N$ in the visible bands $\mathbf{T}^v$ of $\mathbf{T}$.

The overall loss is thus 
\begin{equation}
    L(\mathcal{A}) = \underbrace{\Psi(\mathcal{A},\mathcal{D})}_{\textrm{cloud sensitive}} + \alpha\cdot \underbrace{\Phi(\mathcal{A}, \mathbf{C})}_{\textrm{multispectral}} + \beta \cdot \underbrace{\Omega(\mathcal{A}, \mathbf{T})}_{\textrm{visible domain}}, \label{eq:overall_loss}
\end{equation}
where weights $\alpha, \beta \ge 0$ control the relative importance of the terms. Notice that the loss incorporates multiple objectives across different parts of the spectrum.

\vspace{-1em}
\paragraph{Optimisation}

Minimising $L$ with respect to $\mathcal{A}$ is achieved using the Adam~\cite{kingma2014adam} stochastic optimisation algorithm. Note that the pre-trained cloud detector $f$ is not updated.

\vspace{-1em}
\paragraph{Parameter settings}

See Sec.~\ref{sec:results}.

\subsection{Limitations on real-world testing}\label{sec:limitations}

While our adversarial cube is optimised to be physically realisable, two major constraints prevent physical testing:
\begin{itemize}[leftmargin=1em,itemsep=2pt,parsep=0pt,topsep=2pt]
\item Lack of precise knowledge of and control over the operation of a real satellite makes it difficult to perform coordinated EO data capture with the adversarial cube.
\item Cube dimensions of about 100$\times$100 pixels are required for effective attacks, which translates to 2 km$\times$2 km = 4 km$^2$ ground size (based on the ground resolution of the data; see Sec.~\ref{sec:data_collection}). This prevents full scale fabrication on an academic budget. However, the size of the cube is well within the realm of possibility, \eg, solar farms and airports can be much larger than $4$ km$^2$~\cite{ong2013land}.
\end{itemize}
We thus focus on evaluating our attack in the digital domain, with real-world testing left as future work.

\section{Measuring effectiveness of attacks}\label{sec:metrics}

Let $\mathbf{P}^\ast = \mathbf{P}(\mathcal{A}^\ast)$ be the adversarial cube optimised by our method (Sec.~\ref{sec:attacking}). Recall from Sec.~\ref{sec:data_collection} that both datasets $\mathcal{D}$ and $\mathcal{E}$ contain \emph{non-cloudy} data cubes. We measure the effectiveness of $\mathbf{P}^\ast$ on the training set $\mathcal{D}$ via two metrics:
\begin{itemize}[leftmargin=1em,itemsep=2pt,parsep=0pt,topsep=2pt]
\item Detection accuracy of the pretrained cloud detector $f$ (Sec.~\ref{sec:training}) on $\mathcal{D}$ embedded with $\mathbf{P}^\ast$, i.e.,
\begin{equation}\label{eq:accuracy}
    \text{Accuracy}({\mathcal{D}}) \triangleq 
    \frac{1}{|\mathcal{D}|} 
    \sum^{|\mathcal{D}|}_{k=1} \mathbb{I}(f(\mathbf{D}^c_k(\mathcal{A}^\ast)) \le 0.5),
\end{equation}
where the lower the accuracy, the less often $f$ predicted the correct class label (\emph{non-cloudy}, based on confidence threshold $0.5$), hence the more effective the $\mathbf{P}^\ast$.

\item Average confidence of the pretrained cloud detector $f$ (Sec.~\ref{sec:training}) on $\mathcal{D}$ embedded with $\mathbf{P}^\ast$, i.e.,
\begin{equation}\label{eq:average_probability}
    \text{Cloudy}({\mathcal{D}}) \triangleq 
    \frac{1}{|\mathcal{D}|} 
    \sum^{|\mathcal{D}|}_{k=1} f(\mathbf{D}^c_k(\mathcal{A}^\ast).
\end{equation}
The higher the avg confidence, the more effective the $\mathbf{P}^\ast$.
\end{itemize}
To obtain the effectiveness measures on the testing set $\mathcal{E}$, simply swap $\mathcal{D}$ in the above with $\mathcal{E}$.


\section{Results}\label{sec:results}

We optimised adversarial cubes of size 100$\times$100 pixels on $\mathcal{D}$ (512$\times$512 pixel dimension) under different loss configurations and evaluated them digitally (see Sec.~\ref{sec:limitations} on obstacles to real-world testing). Then, we investigated different cube designs and mitigation strategies for our attack. 

\subsection{Ablation tests}\label{sec:ablation}

Based on the data collected, we optimised adversarial cubes under different combinations of loss terms:
\begin{itemize}[leftmargin=1em,itemsep=2pt,parsep=0pt,topsep=2pt]
    \item $\Psi$: Adversarial biasing in the cloud-sensitive bands~\eqref{eq:loss}.
    \item $\Phi$: Multispectral NPS~\eqref{eq:nps_loss}.
    \item $\Omega$-Hills: Cloaking~\eqref{eq:cloaking_loss} with $\mathbf{T}$ as Hills (Fig.~\ref{fig:hills}).
    \item $\Omega$-Desert: Cloaking~\eqref{eq:cloaking_loss} with $\mathbf{T}$ as Desert (Fig.~\ref{fig:desert}).
\end{itemize}
The weights in~\eqref{eq:overall_loss} were empirically determined to be $\alpha = 5.0$ and $\beta = 0.05$.

\vspace{-1em}
\paragraph{Convex hull and NPS}

\begin{figure*}[ht]\centering
     \begin{subfigure}[b]{0.33\textwidth}
         \includegraphics[width=\textwidth]{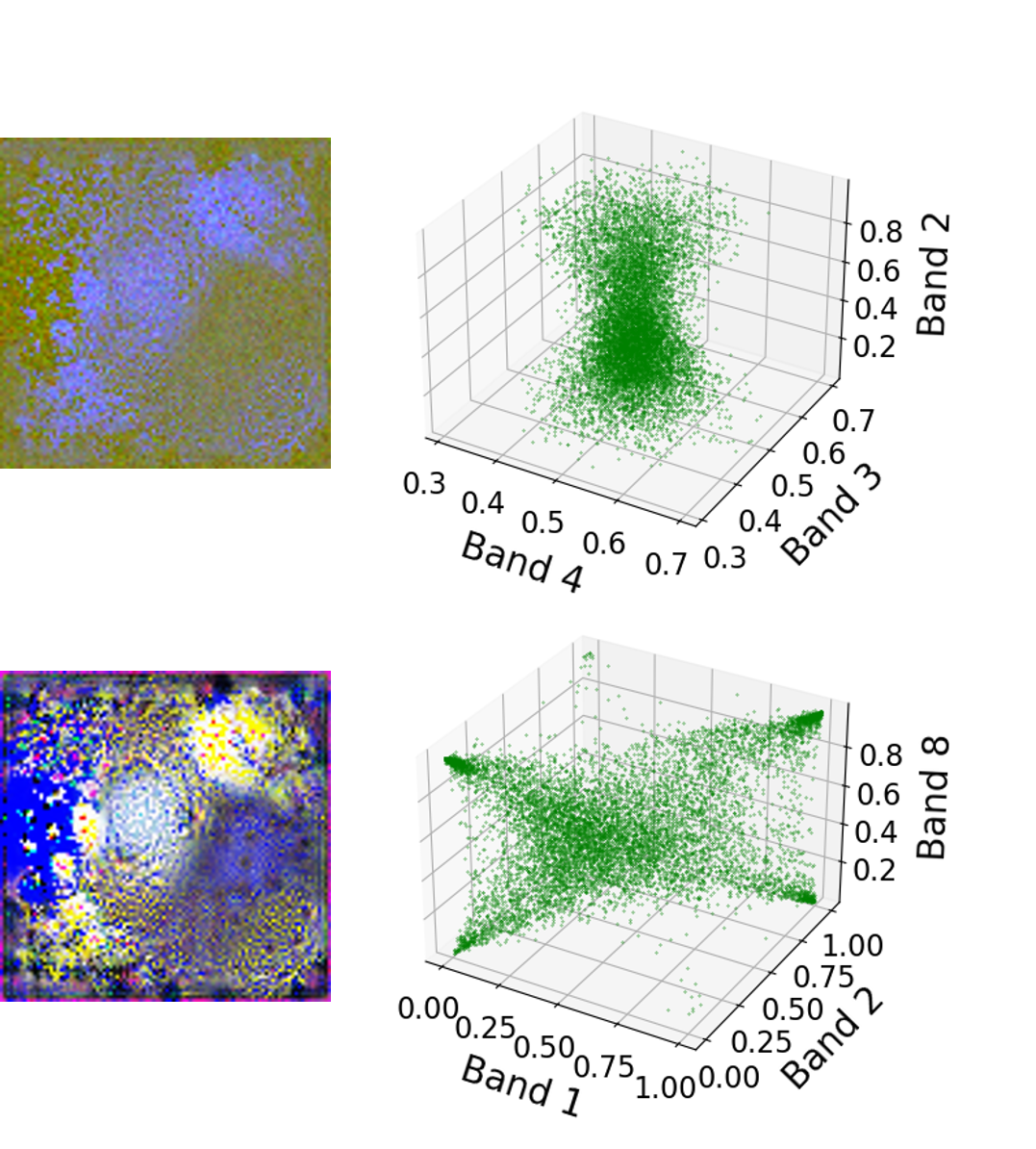}
         \caption{$L = \Psi$ (without convex hull constraints).}
     \end{subfigure}
     \begin{subfigure}[b]{0.33\textwidth}
         \includegraphics[width=\textwidth]{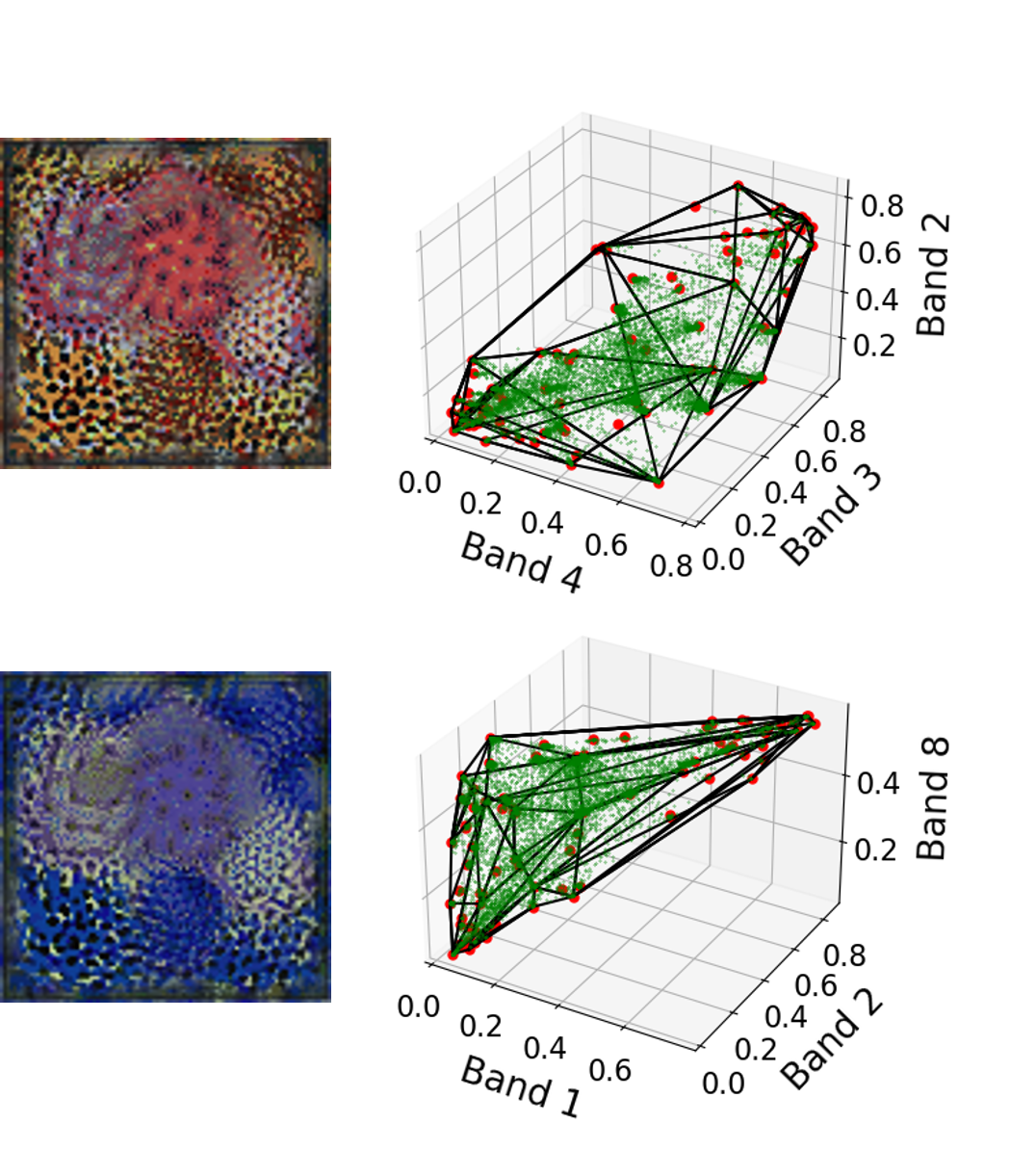}
         \caption{$L = \Psi$.}
     \end{subfigure}
     \begin{subfigure}[b]{0.33\textwidth}
         \includegraphics[width=\textwidth]{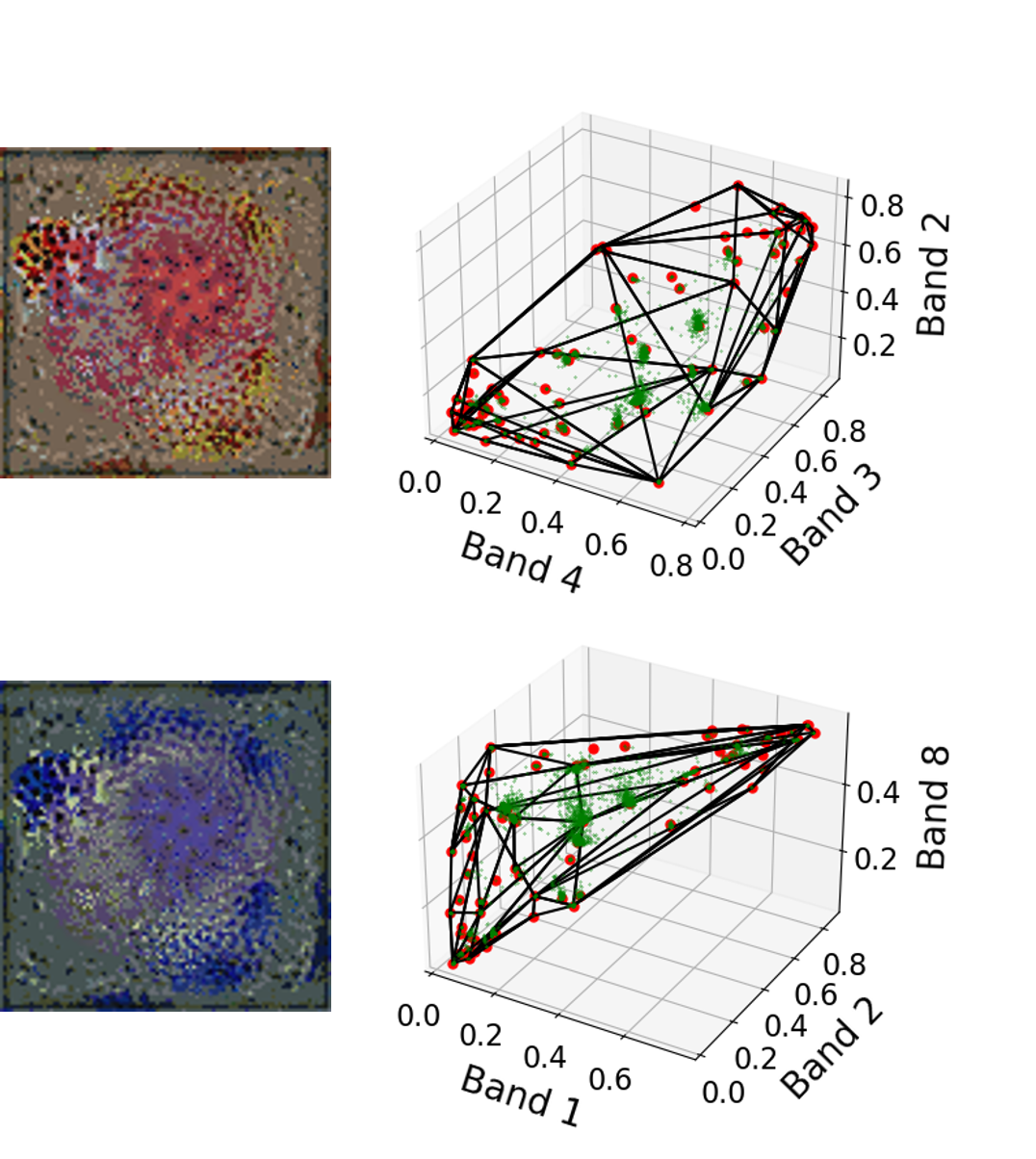}
         \caption{$L = \Psi + \alpha \cdot \Phi$.}
     \end{subfigure}
     \vspace{-0.5em}
     \caption{Effects of convex hull constraints and multispectral NPS on optimised cube $\mathbf{P}^\ast$. The top row shows the cube and individual pixels $\mathbf{p}^\ast_{i,j}$ (green points) in the visible bands $v$, while the bottom row shows the equivalent values in the cloud sensitive bands $c$ (in false colour). In the 3-dimensional plots, the red points indicate the columns of the spectral index $\mathbf{C}$ and black lines its convex hull.}
     \label{fig:cubes_hull}
\end{figure*}

Fig.~\ref{fig:cubes_hull} shows the optimised cubes $\mathbf{P}^\ast$ and its individual spectral intensities $\mathbf{p}^\ast_{i,j}$ in the cloud sensitive bands (false colour) and visible domain. Note that without the convex hull constraints, the intensities (green points) are scattered quite uniformly, which complicates physical realisability of the paint mixtures. The convex hull constraints predictably limit the mixtures to be in the convex hull of $\mathbf{C}$. Carath{\'e}odory's Theorem~\cite{caratheodory-theorem} ensures that each $\mathbf{p}^\ast_{i,j}$ can be obtained by mixing at most 13 exterior paints. In addition, the multispectral NPS term encourages the mixtures to cluster closely around the columns of $\mathbf{C}$ (red points), \ie, close to an existing exterior paint colour.




\vspace{-1em}
\paragraph{Visual camouflage}

\begin{figure}[ht]
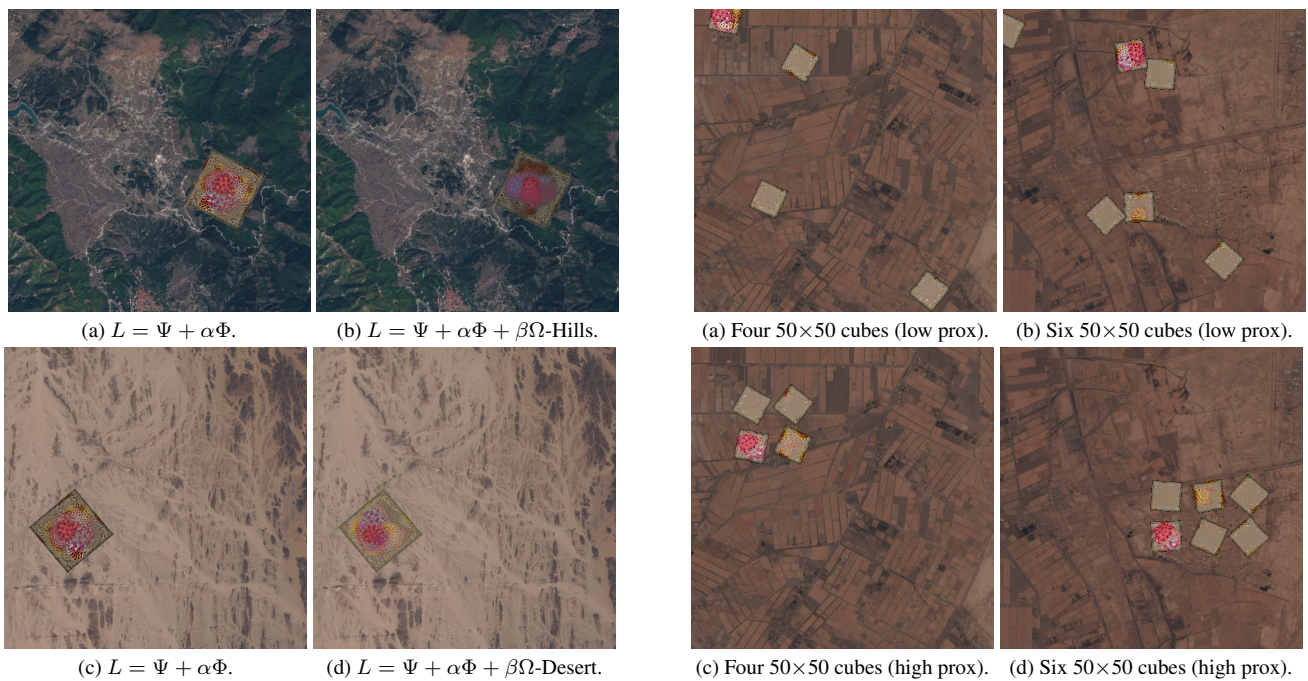
\centering
     \begin{subfigure}[b]{0.23\textwidth}
         \includegraphics[width=\textwidth]{./figures/results/loss/not_camo_hills.pdf}
         \caption{$L = \Psi + \alpha \Phi$.}
     \end{subfigure}
     \begin{subfigure}[b]{0.23\textwidth}
         \includegraphics[width=\textwidth]{./figures/results/loss/camo_hills.pdf}
         \caption{$L = \Psi + \alpha \Phi + \beta \Omega$-$\textrm{Hills}$.}
     \end{subfigure}
     \begin{subfigure}[b]{0.23\textwidth}
         \includegraphics[width=\textwidth]{./figures/results/loss/not_camo_desert.pdf}
         \caption{$L = \Psi + \alpha \Phi$.}
     \end{subfigure}
     \begin{subfigure}[b]{0.23\textwidth}
         \includegraphics[width=\textwidth]{./figures/results/loss/camo_desert.pdf}
         \caption{$L = \Psi + \alpha \Phi + \beta \Omega$-$\textrm{Desert}$.}
     \end{subfigure}     
%
        \vspace{-0.5em}
        \caption{Optimised cubes $\mathbf{P}^\ast$ shown in the visible domain $v$ with and without the cloaking term~\eqref{eq:cloaking_loss}.}
        \label{fig:cubes_loss_images}
\end{figure}

Fig.~\ref{fig:cubes_loss_images} illustrates optimised cubes $\mathbf{P}^\ast$ embedded in the ROA Hills and Desert, with and without including the cloaking term~\eqref{eq:cloaking_loss} in the loss function. Evidently the cubes optimised with $\Omega$ are less perceptible.

\vspace{-1em}
\paragraph{Effectiveness of attacks}

Table~\ref{tab:result_loss} shows quantitative results on attack effectiveness (in terms of the metrics in Sec.~\ref{sec:metrics}) on the training $\mathcal{D}$ and testing $\mathcal{E}$ sets---again, recall that these datasets contain only \emph{non-cloudy} data cubes. The results show that the optimised cubes are able to strongly bias the pretrained cloud detector, by lowering the accuracy by at least $63\%$ (1.00 to 0.37) and increasing the cloud confidence by more than $1000\%$ (0.05 to 0.61). The figures also indicate the compromise an attacker would need to make between the effectiveness of the attack, physical realisablity and visual imperceptibility of the cube.

\begin{table}[ht]
    \setlength\tabcolsep{1pt}
    \centering
    \begin{tabular}{p{4.0cm} | p{1.0cm} p{1.0cm} | p{1.0cm} p{1.0cm}}
    \rowcolor{black} & \multicolumn{2}{l |}{\textcolor{white}{\textbf{Accuracy}}} & \multicolumn{2}{l}{\textcolor{white}{\textbf{Cloudy}}} \\
    \hline
    \textbf{Loss functions} & \textbf{Train} & \textbf{Test} & \textbf{Train} & \textbf{Test} \\
    \hline
    - (no adv.~cubes)       & 1.00      & 1.00              & 0.05      & 0.05 \\ 
    $\Psi$ (no convex hull constr.)        & 0.04      & 0.03              & 0.95      & 0.95 \\ 
    $\Psi$                              & 0.13      & 0.12              & 0.81      & 0.83 \\ 
    $\Psi + \alpha\Phi$                 & 0.22      & 0.19              & 0.73      & 0.75 \\ 
    $\Psi + \beta\Omega$-Hills          & 0.17      & 0.14              & 0.77      & 0.80 \\ 
    $\Psi + \beta\Omega$-Desert         & 0.23      & 0.25              & 0.72      & 0.73 \\ 
    $\Psi + \alpha\Phi + \beta\Omega$-Hills    & 0.25      & 0.28              & 0.71      & 0.70 \\ 
    $\Psi + \alpha\Phi + \beta\Omega$-Desert   & 0.37      & 0.37              & 0.61      & 0.61 \\ 
    \end{tabular}
    \vspace{-0.5em}
    \caption{Effectiveness of 100$\times$100 adversarial cubes optimised under different loss configurations (Sec.~\ref{sec:ablation}). Lower accuracy = more effective attack. Higher cloud confidence = more effective attack.}
    \label{tab:result_loss}
\end{table}

\subsection{Different cube configurations}\label{sec:multcube}

Can the physical footprint of the adversarial cube be reduced to facilitate real-world testing? To answer this question, we resize $\mathbf{P}$ to 50$\times$50 pixels and optimise a number of them (4 or 6) instead. We also tested random configurations with low and high proximity amongst the cubes. The training pipeline for the multi-cube setting remains largely the same. Fig.~\ref{fig:cubes_config_images} shows (in visible domain) the optimised resized cubes embedded in a testing data cube.

Quantitative results on the effectiveness of the attacks are given in Table~\ref{tab:result_cubeconfig}. Unfortunately, the results show a significant drop in attack effectiveness when compared against the 100$\times$100 cube on all loss configurations. This suggests that the size and spatial continuity of the adversarial cube are important factors to the attack.

\begin{table}[ht]
    \setlength\tabcolsep{1pt}
    \centering
    \begin{tabular}{p{0.7cm} | p{1.50cm} | p{1.80cm} | p{1.0cm} p{1.0cm} | p{1.0cm} p{1.0cm}}
    \rowcolor{black} \multicolumn{3}{l |}{\textcolor{white}{\textbf{Cube configurations}}} & \multicolumn{2}{l |}{\textcolor{white}{\textbf{Accuracy}}} & \multicolumn{2}{l}{\textcolor{white}{\textbf{Cloudy}}} \\
    \hline
    \textbf{\#} & \textbf{Size} & \textbf{Proximity} & \textbf{Train} & \textbf{Test} & \textbf{Train} & \textbf{Test} \\
    \hline
    \multicolumn{3}{l |}{- (no adv.~cubes)} & 1.00      & 1.00      & 0.05      & 0.05 \\ 
    4 & 50$\times$50         & Low      & 0.87      & 0.87      & 0.26      & 0.27 \\ %
    6 & 50$\times$50         & Low      & 0.71      & 0.72      & 0.33      & 0.33 \\ %
    4 & 50$\times$50         & High     & 0.63      & 0.62      & 0.42      & 0.44 \\
    6 & 50$\times$50         & High     & 0.63      & 0.63      & 0.40      & 0.41 \\
  \end{tabular}
  \vspace{-0.5em}
  \caption{Effectiveness of 50$\times$50 adversarial cubes under different cube configurations (Sec.~\ref{sec:multcube}) optimised with loss $L = \Psi + \alpha\Phi$. Lower accuracy = more effective attack. Higher cloud confidence = more effective attack. Compare with single 100$\times$100 adversarial cube results in Table~\ref{tab:result_loss}.}
  \label{tab:result_cubeconfig}
\end{table}

\begin{figure}[ht]\centering
     \begin{subfigure}[b]{0.23\textwidth}
         \includegraphics[width=\textwidth]{./figures/results/config/four_random.pdf}
         \caption{Four 50$\times$50 cubes (low prox).}
     \end{subfigure}
     \begin{subfigure}[b]{0.23\textwidth}
         \includegraphics[width=\textwidth]{./figures/results/config/six_random.pdf}
         \caption{Six 50$\times$50 cubes (low prox).}
     \end{subfigure}
     \begin{subfigure}[b]{0.23\textwidth}
         \includegraphics[width=\textwidth]{./figures/results/config/four_fixed.pdf}
         \caption{Four 50$\times$50 cubes (high prox).}
     \end{subfigure}
     \begin{subfigure}[b]{0.23\textwidth}
         \includegraphics[width=\textwidth]{./figures/results/config/six_fixed.pdf}
         \caption{Six 50$\times$50 cubes (high prox).}
     \end{subfigure}
        \vspace{-0.5em}
        \caption{Optimised cubes $\mathbf{P}^\ast$ shown in the visible domain $v$ of different cube configurations.}
        \label{fig:cubes_config_images}
\end{figure}

\subsection{Mitigation strategies}\label{sec:mitigation}

We investigated several mitigation strategies against our adversarial attack:
\begin{itemize}[leftmargin=1em,itemsep=2pt,parsep=0pt,topsep=2pt]
    \item 13 bands: Increasing the number of input bands of the cloud detector from 3 to 13 (all Sentinel-2A bands); 
    \item $\sqrt{2}$: Doubling the model size of the cloud detector by increasing the number of filter/kernels in the convolutional layers and activations in the fully connected layers by $\sqrt{2}$ 
    \item $2\times$ CONV: Doubling the model size of the cloud detector by adding two additional convolutional layers.  
\end{itemize}
Table~\ref{tab:result_mitigations} shows that using a ``larger'' detector (in terms of the number of input channels and layers) yielded slightly worse cloud detection accuracy. However, increasing the number of input bands significantly reduced our attack effectiveness, possibly due to the increased difficulty of biasing all 13 channels simultaneously. This argues for using greater satellite-borne compute payloads than that of~\cite{giuffrida2020cloudscout}.

\begin{table}[ht]
    \setlength\tabcolsep{1pt}
    \centering
    \begin{tabular}{p{1.5cm} | p{2.5cm} | p{1.0cm} p{1.0cm} | p{1.0cm} p{1.0cm}}
    \rowcolor{black} &   & \multicolumn{2}{l |} {\textcolor{white}{\textbf{Accuracy}}}  & \multicolumn{2}{l} {\textcolor{white}{\textbf{Cloudy}}} \\
    \hline
    \textbf{Detectors} & \textbf{Loss functions} & \textbf{Train} & \textbf{Test} & \textbf{Train} & \textbf{Test} \\
    \hline
    13 bands        & - (no adv.~cubes)   & 1.00      & 1.00      & 0.06      & 0.06 \\ 
                    & $\Psi + \alpha\Phi$   & 0.94      & 0.96      & 0.15      & 0.14 \\ 
    \hline
    $\sqrt{2}$      & - (no adv.~cubes)   & 1.00     & 1.00      & 0.08      & 0.08 \\ 
                    & $\Psi + \alpha\Phi$   & 0.36      & 0.38      & 0.62      & 0.60 \\    
    \hline
    $2\times$CONV   & - (no adv.~cubes)   & 1.00     & 1.00      & 0.08      & 0.08 \\
                    & $\Psi + \alpha\Phi$   & 0.26      & 0.25      & 0.74      & 0.73 \\    
    \end{tabular}
    \vspace{-0.75em}
    \caption{Effectiveness of 100$\times$100 adversarial cubes optimised for different cloud detector designs (Sec.~\ref{sec:mitigation}). Lower accuracy = more effective attack. Higher cloud confidence = more effective attack. Compare with single 100$\times$100 adversarial cube results in Table~\ref{tab:result_loss}.}
    \label{tab:result_mitigations}
\end{table}

\section{Conclusions and limitations}\label{sec:conclusion}

We proposed a physical adversarial attack against a satellite-borne multispectral cloud detector. Our attack is based on optimising exterior paint mixtures that exhibit the required spectral signatures to bias the cloud detector. Evaluation in the digital domain illustrates the realistic threat of the attack, though the simple mitigation strategy of using all input multispectral bands seems to offer good protection.

As detailed in Sec.~\ref{sec:limitations}, our work is limited to digital evaluation due to several obstacles. Real-world testing of our attack and defence strategies will be left as future work.

\vfill

\section{Usage of existing assets and code release}

The results in this paper were partly produced from ESA remote sensing data, as accessed through the Copernicus Open Access Hub~\cite{2021copernicus}. Source code and/or data used in our paper will be released subject to securing permission.


\vfill

\section*{Acknowledgements}\label{sec:acknowledgement}
Tat-Jun Chin is SmartSat CRC Professorial Chair of Sentient Satellites.



{\small
\bibliographystyle{ieee_fullname}
\bibliography{egbib}
}

\clearpage

\section{Supplementary Material}\label{sec:supp}

Additional information about the datasets used for training, validating and testing the cloud detector as well as the CNN architecture of Giuffrida \etal and training parameters \cite{giuffrida2020cloudscout} are provided below. 

\subsection{TH30 and TH70 datasets}\label{sec:cmc_supp}
Table~\ref{tab:cm_dataset} provides a summary of the datasets derived from the Cloud Mask Catalogue~\cite{francis_alistair_2020_4172871} used to train, validate and test the cloud detector.

\begin{table}[ht]
    \setlength\tabcolsep{1pt}
        \centering
        \begin{tabular}{p{3.5cm} | p{2.25cm} | p{2.25cm}}
        \rowcolor{black}    & \textcolor{white}{\textbf{TH30}} & \textcolor{white}{\textbf{TH70}} \\
        \hline 
        \textbf{Training}           & \\            
        \hline
        Number of data cubes        & 1,232                     & 1,356 \\
        Data augmentations          & Horizontal flip           & Horizontal flip\\
                                    & Vertical flip             & Vertical flip \\
                                    & Gaussian noise            & Gaussian noise \\
        \hline
        \textbf{Validation}         & \\
        \hline
        Number of data cubes        & 264                       & 290 \\
        Data augmentations          & Horizontal flip           & Horizontal flip \\
        \hline
        \textbf{Testing}            &   \\
        \hline
        Number of data cubes        & 556                       & 406 \\
        Data augmentations          & -                         & - \\
        \hline
        \end{tabular}
     \vspace{-0.25em}    
     \caption{Details of the TH30 and TH70 datasets.}
     \label{tab:cm_dataset}
\end{table}


\subsection{CNN architecture of the cloud detector}\label{sec:model_supp}
\begin{figure}[ht]\centering
	 \includegraphics[width=1.0\linewidth]{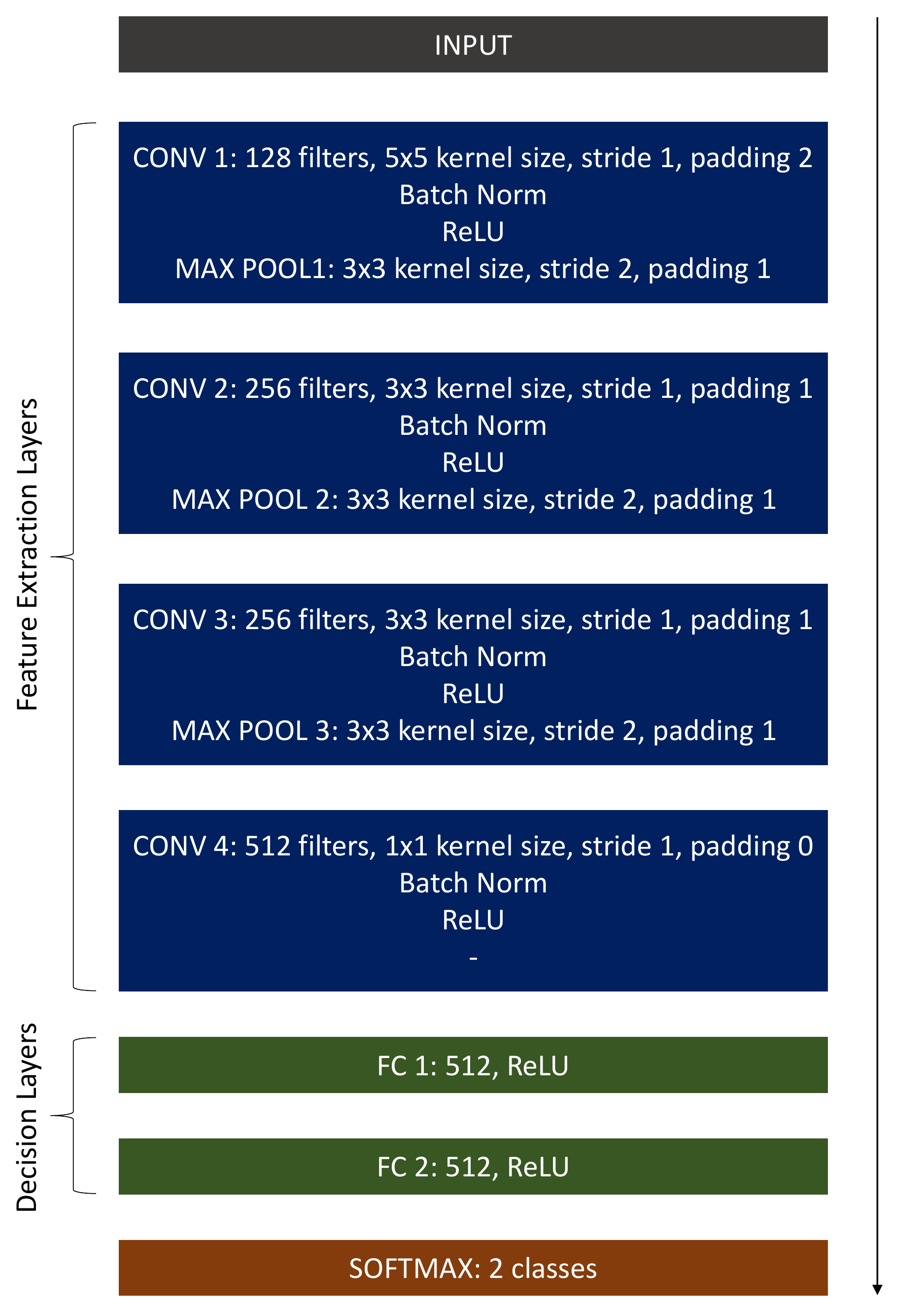}
     \vspace{-1.5em}
	 \caption{CNN architecture of Giuffrida \etal. \cite{giuffrida2020cloudscout}.}
	 \label{fig:cnn_model}
\end{figure}

Fig.~\ref{fig:cnn_model} shows the CNN architecture and model hyperparameters of the cloud detector.

\subsection{Training parameters of the cloud detector}

The training parameters used in \cite{giuffrida2020cloudscout} are provided here. We used an initial learning rate of $\eta_{0} = 0.01$ and exponentially decayed that rate by computing:
\begin{equation}
    \eta_{k+1} = \eta_{k} \cdot \exp(-0.6 \cdot k), 
    \label{eq:learning_rate}
\end{equation}
at every $k$ epoch. The Binary-Cross Entropy loss function was modified by doubling the weight term for false positive error cases to trade-off false positives for false negatives:
\begin{equation}
    L(y,\hat{y}) = - y \cdot \log(\hat{y}) - 2 \cdot (1 - y) \cdot \log(1 - \hat{y}), 
    \label{eq:cross_entropy}
\end{equation}
where $y$ is the ground truth and $\hat{y}$ is the predicted probability of the \textit{cloudy} class. The cloud detector was trained on TF30 for 300 epochs and fine-tuned on TF70 for 300 epochs.

\end{document}